%% file: main.tex
\documentclass{article}

\usepackage{cite}
\usepackage{amsmath,amssymb,amsfonts}
\usepackage{algorithmic}
\usepackage{graphicx}
\usepackage{textcomp}
\usepackage{xcolor}
\def\BibTeX{{\rm B\kern-.05em{\sc i\kern-.025em b}\kern-.08em
    T\kern-.1667em\lower.7ex\hbox{E}\kern-.125emX}}

\newcommand{\cut}[1]{}

\DeclareMathAlphabet{\pazocal}{OMS}{zplm}{m}{n}

\begin{document}

\title{Challenges and Opportunities in Approximate Bayesian Deep Learning for Intelligent IoT Systems
% {\footnotesize \textsuperscript{*}Note: Sub-titles are not captured in Xplore and
% should not be used}
}

% \author{\IEEEauthorblockN{Meet P. Vadera}\\
% \IEEEauthorblockA{\textit{Manning College of Information and Computer Sciences} \\
% \textit{University of Massachusetts Amherst}\\
% Amherst, MA, USA \\
% mvadera@cs.umass.edu}
% \and
% \IEEEauthorblockN{Benjamin M. Marlin}\\
% \IEEEauthorblockA{\textit{Manning College of Information and Computer Sciences} \\
% \textit{University of Massachusetts Amherst}\\
% Amherst, MA, USA \\
% marlin@cs.umass.edu}
% }
\author{Meet P. Vadera, and Benjamin M. Marlin\\
\textit{Manning College of Information and Computer Sciences} \\
\textit{University of Massachusetts Amherst}\\
Amherst, MA, USA \\
\{mvadera, marlin\}@cs.umass.edu
}
\date{}
\maketitle

\input{abstract}

% \begin{IEEEkeywords}
% Bayesian deep learning, Internet of Things, Deep Learning, Model Compression.
% \end{IEEEkeywords}

\input{introduction}
\input{bnn_background}
\input{model_compression}
\input{distillation}
%\input{evaluation}
%\input{opportunities_challenges}
\input{conclusions}
\input{acknowledgement}

\bibliography{references}
\bibliographystyle{IEEEtran}

\end{document}

%% file: abstract.tex
\begin{abstract}
Approximate Bayesian deep learning methods hold significant promise for addressing several issues that occur when deploying deep learning components in intelligent systems, including mitigating the occurrence of over-confident errors and providing enhanced robustness to out of distribution examples. However, the computational requirements of existing approximate Bayesian inference methods can make them ill-suited for deployment in intelligent IoT systems that include lower-powered edge devices. In this paper, we present a range of approximate Bayesian inference methods for supervised deep learning and highlight the challenges and opportunities when applying these methods on current edge hardware. We highlight several potential solutions to decreasing model storage requirements and improving computational scalability, including model pruning and distillation methods. 
\end{abstract}

%% file: introduction.tex
\section{Introduction}
\label{sec:intro}

Deep learning research has shown promising results in many application areas of artificial intelligence including object detection, language modeling, speech recognition, medical imagining, image segmentation and many more \cite{graves2013speech,Huang2016DenselyCC,Devlin2018BERTPO, Singh2016MachineLF, Singh2019MulticlassDO, suzuki2017overview, DBLP:conf/chase/VaderaV19, DBLP:journals/corr/abs-2001-05566, Vadera2021post}. Key components driving the overall success of deep learning-based methods include advances in learning algorithms, neural network architectures, computing hardware including graphics processing units (GPUs) and tensor processing units (TPUs), and the availability of large labeled data sets.

%However, there remains a fundamental issue with how deep learning methods have been used traditionally. 

However, as deep learning models are being deployed in fielded intelligent systems, including intelligent IoT systems, several challenges have become increasingly prominent. These challenges include the deployment-time occurrence of high confidence errors, the need to be robust to out-of-distribution inputs, and the potential for in-domain adversarial inputs. High confidence errors occur when a probabilistic machine learning model ascribes high probability to an incorrect output (e.g., a class label in a classification setting) \cite{guo2017calibration}. Deployed deep learning models can also encounter inputs from data distribution that differ systematically from the distribution they were trained on. In such cases, models need to avoid making high confidence errors. One potential approach to dealing with this issue is to explicitly identify inputs as out-of-distribution and to decline to make predictions for them \cite{Hendrycks2017ABF}. Lastly, intelligent systems can also encounter adversarial inputs of different types. Early work in adversarial example generation focused on algorithms for making low-norm changes to inputs that, while being nearly imperceptible to humans, result in models making highly confident errors \cite{Goodfellow2015ICLR}.  

One of the important factors contributing to susceptibility to these problems is model uncertainty. Supervised deep learning models are most commonly trained by optimizing their parameters to minimize a training loss function. This approach yields a single locally optimal setting of the model parameters, which are then used to make predictions at deployment time. However, given the large number of parameters used in current models, there typically exist multiple qualitatively different sets of parameters yielding similar training loss function values, but making different predictions on future inputs. Commmon training procedures select a single set of such parameters, despite the fact that for a given amount of training data, there may be significant uncertainty over what the best model parameters actually are. 

%exists multiple settings of parameters that can fit well to the training data distribution. Each of the parameter settings possible have their own unique bias as we extrapolate away from training data, resulting in overconfident incorrect predictions at times. Bayesian Neural Networks (BNNs) can help mitigate these issues by marginalizing over the parameter posterior distribution.

Bayesian inference provides a different perspective on the problem of training deep neural network models that attempts to represent model uncertainty and propagate it to the point of issuing predictions and making decisions. In Bayesian inference, the unknown model parameters are formally treated as random variables. The goal becomes to infer the posterior probability distribution over the unknown model parameters given the available data. When the data support multiple distinct interpretations in terms of settings of the model parameters, the model posterior will reflect this uncertainty. Incorporating model uncertainty into prediction and decision-making typically decreases overconfidence in predictions via the Bayesian model averaging effect, and can also increase robustness to out-of-distribution and adversarial inputs \cite{wilson2020case, Vadera2020AssessingTA}. Quantities computed from the model posterior can also be used as inputs for auxiliary problems, including the detection of out-of-distribution examples.     

%As we continue to build intelligent IoT systems for practical purposes, Bayesian deep learning methods become a focal point of interest. Intelligent IoT systems often combine multiple data modalities \cite{DBLP:conf/cogmi/MarlinACCDJKRRS20}, and due to the high stakes involved, require reliable outputs. Additional desirable properties involve detecting out-of-distribution inputs, adversarial inputs in-domain distribution inputs that the model is uncertain about, and wants to defer the decision to humans. Indeed, Bayesian neural networks help with each of these aspects, as we touch upon related uncertainty decomposition \cite{Depeweg2017DecompositionOU}, Bayesian active learning \cite{houlsby2011bayesian, DBLP:conf/nips/KirschAG19, conrad2019towards}, and Bayesian decision theory \cite{Berger1988StatisticalDT, lacoste2011approximate, Cobb2018LossCalibratedAI, kusmierczyk2019variational, Vadera2021post}.

Bayesian inference has the potential to provide a comprehensive theoretical foundation for constructing uncertainty-aware and robust  deep learning-based intelligent IoT systems, including systems that must reason robustly over complex multi-modal inputs in the field. However, the application of Bayesian inference methods to deep neural network models is challenging due to the large scale of current state-of-the-art prediction models. 

In this paper, we focus on the challenges and opportunities that arise when we consider deploying Bayesian deep learning approaches on IoT edge devices.
Specifically, traditional approximate Bayesian learning algorithms represent the model posterior as a large ensemble of models. This can be expensive from the storage perspective, as the cost of storing an ensemble of $S$ models will be $S$ times higher than the cost of storing a single model unless additional compression is used. Further, when making predictions, traditional Bayesian ensembles require processing each input instance through each element of the ensemble. Again, this requires $S$ times more compute when using an ensemble of $S$ models compared to the use of a single model. 

The remainder of this paper is organized as follows. In Section \ref{sec:bnn_background} we begin by providing a comprehensive discussion of Bayesian supervised learning, approximate Bayesian inference, and the scalability challenges of deploying current Bayesian deep learning model representations on edge hardware.  Next, in sections \ref{sec:model_compression_pruning} and \ref{sec:model_compression_distillation} we discuss model compression techniques that can be leveraged for compressing Bayesian posterior distributions. These approaches either look at compressing each member of the model ensemble, or compress the entire ensemble into a surrogate model. We point out challenges, opportunities and open research directions related to both approaches.
% In section \ref{sec:evaluation_considerations} we discuss evaluation criteria that practitioners should consider when evaluating the selection of approximate Bayesian inference and model compression techniques.
%Lastly, in section \ref{sec:opportunities_challenges} we discuss challenges and opportunities that arise when combining the different approaches discussed in this paper to develop intelligent IoT systems.

%% file: bnn_background.tex
\section{Bayesian Deep Learning \\and the Challenge of Scalability}
\label{sec:bnn_background}
In this section we introduce the fundamental concepts of Bayesian inference for supervised deep learning along with foundational approximation methods. We discuss the scalability challenges when deploying such methods on edge and IoT systems.

\subsection{Bayesian Supervised Learning}  
Supervised learning forms the core of machine learning-based prediction systems. In a supervised learning problem, we are given a dataset $\mathcal{D}$ consisting of input-output pairs $\{(\mathbf{x}_i,y_i)|1\leq i\leq N\}$, where $\mathbf{x}_i\in\mathbb{R}^D$ is the input or feature vector and $y_i\in\mathcal{Y}$ is the output or prediction target. We let $\mathcal{D}^x$ be the dataset of inputs and $\mathcal{D}^y$ be the dataset of outputs. The nature of $\mathcal{Y}$ depends on the task at hand. For classification tasks, $\mathcal{Y}$ is a finite set, whereas for regression tasks, we usually have $\mathcal{Y} \in \mathbb{R}$. In this paper, we specifically focus on the classification setting in supervised learning, where the goal is to learn a function $f: \mathbb{R}^D \rightarrow \mathcal{Y}$ that can accurately predict the outputs from the inputs.

In probabilistic supervised learning, we construct the prediction function using a conditional probability model of the form $f_\theta (\mathbf{x}) = p(y|\mathbf{x}, \theta)$ where $\theta\in\mathbb{R}^K$ are the model parameters. The conditional likelihood of the inputs given the outputs and the parameters is denoted by $p(\mathcal{D}^y|\mathcal{D}^x,\theta)$. Under the assumption that the outputs are independent and identically distributed given their corresponding outputs, we have that $p(\mathcal{D}^y|\mathcal{D}^x,\theta)=\prod_{i=1}^N p(y_i|\mathbf{x}_i, \theta)$. A key ingredient in Bayesian inference, as well as in traditional point-estimated neural networks, is the prior distribution $p(\theta|\lambda)$ over model parameters. As the name suggests, the prior distribution represents our beliefs about  the distribution of the model parameters prior to analyzing the data. The prior distribution can have its own hyperparameters, here denoted by $\lambda$ \cite{Neal:1996:BLN:525544}. 

Bayesian inference involves the computation of the posterior distribution over the unknown model parameters given a training dataset $\mathcal{D}_{tr}$ and the prior. The parameter posterior is obtained using the Bayes theorem, as shown in Equation \eqref{eq:posterior}. 

\begin{align}
\label{eq:posterior}
    p(\theta|\mathcal{D}_{tr},\lambda)& = \frac{p(\mathcal{D}_{tr}^y|\mathcal{D}_{tr}^x,\theta)p(\theta|\lambda)}{\int p(\mathcal{D}_{tr}^y|\mathcal{D}_{tr}^x,\theta)p(\theta|\lambda) d\theta}
\end{align}

The denominator in the parameter posterior (referred to as the ``evidence" term) is  intractable to compute for most ML models, including for deep neural networks \cite{Neal:1996:BLN:525544}. As a result, the computation of the exact posterior distribution is intractable. However, in practice, the quantity of interest is often not the parameter posterior distribution itself, but rather low-dimensional expectations under the parameter posterior. 

One key posterior expectation in the supervised learning setting is the posterior predictive distribution, which is necessary and sufficient for making maximum probability predictions for outputs given inputs while integrating over the uncertainty in the model parameters. The posterior predictive distribution computation is shown in Equation \eqref{eq:posterior_predictive}.

\begin{align}
\label{eq:posterior_predictive}
p(y| \mathbf{x}, \mathcal{D}_{tr},\lambda) &= 
\mathbb{E}_{p(\theta|\mathcal{D}_{tr},\lambda)}[p(y|\mathbf{x}, \theta)]
\end{align}

Another posterior expectation that is useful in uncertainty quantification is the expected posterior predictive entropy. Posterior predictive entropy (also referred to as the \emph{total uncertainty} of the predictive distribution) can be decomposed into quantities referred to as \emph{expected data uncertainty} and \emph{knowledge uncertainty} \cite{Depeweg2017DecompositionOU}. These three forms of uncertainty are related by the equation shown below:

\begin{align}
\underbrace{\pazocal{H}\left[\mathbb{E}_{p(\theta | \pazocal{D})}\left[p\left(y | \boldsymbol{x}, \theta\right)\right]\right]}_{\text {Total Uncertainty }} &=    
\underbrace{\pazocal{I}\left[y, \theta | \boldsymbol{x}, \pazocal{D}\right]}_{\text {Knowledge Uncertainty }} \notag \\
&+\underbrace{\mathbb{E}_{p(\theta | \pazocal{D})}\left[\pazocal{H}\left[p\left(y | \boldsymbol{x}, \theta\right)\right]\right]}_{\text {Expected Data Uncertainty }}
\end{align}

%Total uncertainty, as the name suggests, measures the total uncertainty in a prediction. Expected data uncertainty measures the average predictive uncertainty over the posterior Knowledge uncertainty corresponds to the conditional mutual information between outputs and model parameters and measures the disagreement between different models in the posterior. 
%However, 
Knowledge uncertainty can be efficiently computed as the difference between total uncertainty and expected data uncertainty, both of which are functions of posterior expectations. Recent work has leveraged these uncertainty estimates to explore a range of down-stream tasks such out-of-distribution detection, misclassification detection, and active learning that rely on uncertainty quantification and decomposition.\cite{wang2018adversarial, malinin2020ensemble, Vadera2020GeneralizedBP, Vadera2020URSABenchCB, houlsby2011bayesian, conrad2019towards, DBLP:conf/nips/KirschAG19}. 
However, all of these posterior expectations are also intractable to compute exactly for deep learning models. We thus next turn to the problem of approximate Bayesian inference methods. 

\subsection{Approximate Bayesian Inference for Supervised Learning} 

As indicated in the previous subsection, posterior expectations including  the posterior distribution needed for Bayesian supervised learning is intractable in its original form. To tackle this problem there is a significant body of work in the area of approximate Bayesian inference techniques. The ultimate goal of these approximation methods is to compute approximate posterior expectations that are close to their theoretical counterparts. Approximate Bayesian methods can be broadly divided into three categories: Markov Chain Monte Carlo (MCMC) methods,  surrogate density  estimation methods, and other approximation methods. We describe each category of methods below and discuss their edge deployment challenges. 

\subsubsection{Markov Chain Monte Carlo Methods} 
MCMC methods provide an approximation to the intractable parameter posterior $p(\theta|\mathcal{D},\lambda)$ via a set of samples drawn for this distribution. MCMC methods simulate a Markov chain that converges to the parameter posterior as its steady state distribution. The simulated states of the Markov chain after convergence correspond to samples from the parameter posterior. Once we collect a set of parameter samples of the desired size, we can approximate expectations with respect to the parameter posterior using empirical averages over the sampled parameter values \cite{smith1993bayesian}. For example, the posterior predictive distribution can be approximated as shown in Equation \eqref{eq:mc_posterior_predictive}. As we can see, computing the Monte Carlo approximation to the posterior predictive distribution is very similar to computing the predictive distribution of a model ensemble. 
%As a useful theoretical property, we note that the Monte Carlo average converges to at the rate of $\mathcal{O}(1/\sqrt{S})$.
%
\begin{align}
\label{eq:mc_posterior_predictive}
p(y| \mathbf{x}, \mathcal{D},\lambda)
&=\mathbb{E}_{p(\theta|\mathcal{D},\lambda)}[p(y|\mathbf{x}, \theta)] \notag \\ 
& \approx \mathbb{E}_{p_{\textrm{MC}}(\theta|\mathcal{D},\lambda)}[p(y|\mathbf{x}, \theta)] \notag\\ 
&=\frac{1}{S}\sum_{s=1}^S p(y|\mathbf{x}, \theta_s)
\end{align}
Examples of classical MCMC methods include the Gibbs sampler \cite{casella1992explaining} and the Metropolis-Hastings sampler \cite{chib1995understanding}. The earliest work on MCMC samplers for neural networks traces back to the application of Hamiltonian Monte Carlo \cite{duane1987hybrid} methods. While a number of MCMC methods have since been developed with improved properties including slice sampling \cite{neal2003slice}, elliptical slice sampling \cite{Murray2010EllipticalSS}, and Riemann manifold sampling methods \cite{girolami2011riemann}, these methods all require using all of the available data when computing the likelihood term needed for posterior inference. Although only linear in the  number of data cases, this can be a highly expensive operation for large data sets and models and can render MCMC methods practically infeasible in scenarios where stochastic gradient descent (SGD) \cite{bottou2010large} can be usefully applied to optimize model parameters. 

However, recent advances in MCMC approaches have enabled the use of SGD-like mini-batch algorithms, greatly extending the range of applicability of MCMC methods. Prominent examples of such approaches include stochastic gradient Langevin dynamics (SGLD) \cite{welling2011bayesian}, stochastic gradient Hamiltonian Monte Carlo (SGHMC) \cite{Chen2014StochasticGH} and their cyclic learning rate versions as presented in \cite{Zhang2020Cyclical}. 

Another important property that determines the efficacy of MCMC methods is the degree of mixing. The degree of mixing refers to how efficiently the Markov chain traverses  the posterior distribution after convergence \cite{brooks2011handbook}. Better mixing enables faster collection of a more diverse set of parameter samples. However, the mixing properties of MCMC methods depend on the dimensionality of the parameter space. Modern deep neural networks can have an extremely large number of parameters (millions or more), potentially leading to inadequate mixing. 

An alternative to sampling in the original parameter space $\mathbb{R}^K$ is to sample in a reduced dimensional space $\mathbb{R}^{K'}$ for $K'\ll K$ and to project back to the full-dimensional space. This process is termed \textit{subspace inference} \cite{Izmailov2019SubspaceIF}. Subspaces can be generated using singular value decomposition (SVD) applied to SGD iterates, or by generating random projection matrices, among other possible options \cite{Izmailov2019SubspaceIF}. These methods  introduce bias by restricting the sampler to operate in a subspace of $\mathbb{R}^K$, but reduce variance by enabling the underlying Markov chain to mix faster.

While recent advances in MCMC methods are enabling the application of Bayesian inference approaches to increasingly large deep learning models, there remains a sizeable gap in terms of the practical deployment of Monte Carlo-based posterior approximations on edge hardware to support intelligent IoT systems. As noted previously, MCMC methods generate $S$ samples from the model posterior in place of the single set of parameters used by optimization-based deep learning. This means that the deployment of Monte Carlo-based approximate prediction methods requires $S$ times more storage as well as $S$ times more computation without further  approximations.

However, the actual increase in prediction latency depends on a number of factors. First, in the edge setting, it is typical for an edge device to process data from on-board sensors. In this case, the edge device may only need to make predictions for a single instance at a time. For GPU or TPU accelerated edge devices, it may thus be possible to run a single input through multiple models with identical structure in parallel, effectively batching the models instead of the inputs. The feasibility of this approach depends on the size of the model to be deployed. Second, many edge platform toolchains have the ability to perform weight quantization and to use mixed precision arithmetic when deploying models. Such transformations can be applied separately to each element of a posterior ensemble to improve edge deployment efficiency. We note that the effect of such quantization has not been well studied for Bayesian posterior ensembles, but there is reason to believe that they might better tolerate more aggressive quantization than single models due to the average that is taken over the outputs of the posterior ensemble when making predictions.  

Lastly, we note that the memory constraints of edge devices can also play a non-trivial role in prediction latency when using ensembles due to the time required to load models. In some instances, the time needed to load models is significantly higher than the time needed to use the loaded model to make predictions for a single input. When this is the case, it can dramatically increase prediction latency. Indeed, the number of models that can be present in memory simultaneously can currently be the effective limiting factor on the size of the MCMC posterior ensemble that can usefully be deployed. 

An interesting opportunity motivated by these observations is the development of optimized model loading methods specifically for ensembles of models with identical architectures. In this case, the structure of the computation graph underlying the architecture should only need to be instantiated once, and it may be possible to more rapidly iterate through different sets of parameters for the same architecture than is possible using current libraries that do not include such optimizations \cite{tensorrt}.

\subsubsection{Surrogate Density Methods}
An alternate approach to MCMC methods is approximating the original posterior distribution via an analytically tractable parameterized surrogate distribution. Thus, given the original posterior distribution $p(\theta|\mathcal{D}_{tr},\lambda)$, surrogate density methods aim to approximate the true posterior using an auxiliary distribution $q(\theta|\phi)$, where $\phi$ are the auxiliary parameters \cite{jordan1999introduction, jaakkola2000bayesian, ghosh2016assumed, minka2001expectation}. A common approximation for auxiliary parameters is the use of a multivariate Gaussian distribution with a diagonal covariance matrix $\mathcal{N}(\theta;\mu,\Sigma)$, also known as  mean-field variational inference. Here, the auxiliary parameters are $\phi=[\mu,\Sigma]$. The main reason why mean-field variational inference is popular is due to the simple ``re-parameterization trick" that makes sampling from the auxiliary distribution straightforward \cite{blundell2015weight}. With the re-parameterization trick, we can sample  $\theta_k \sim \mathcal{N}(\mu_k, \Sigma_{kk})$ by first drawing $\eta \sim \mathcal{N}(0, 1)$, followed by the linear transformation $\theta_k = \mu_k + \eta \cdot \sqrt{\Sigma_{kk}}$. This allows us to backpropagate through the variational parameters while drawing samples of the model parameters to approximate the objective functions used for learning. 

Now, for estimating the auxiliary parameters, we need an objective function that can measure the discrepancy between the surrogate distribution and the ground truth posterior. Needless to say, this objective function must also be computationally tractable. The most commonly used discrepancy function is the Kullback-Leibler (KL) divergence as shown below \cite{mackay2003information}. $$\textrm{KL}(p(\theta)||q(\theta)) = \mathbb{E}_{p(\theta)}\left[\log \left(\frac{p(\theta)}{q(\theta)}\right)\right]$$

The KL divergence is a directional divergence, and thus is not symmetric in its arguments. This can result in two different measures depending on the directionality. When the surrogate posterior is used as the first argument, the result is the variational inference (VI) framework \cite{jordan1999introduction, jaakkola2000bayesian}. When the surrogate posterior is used as the second argument, the result is the expectation propagation (EP) framework \cite{minka2001expectation}. This often leads to VI methods having mode seeking behavior, as they are not forced to ensure the same support as the original posterior. EP methods on the other hand are forced to ensure exact support, but this can often lead to incorrect mode estimation as it must cover the support of the original posterior. These two extremes can also be interpolated by more generalized divergence measures, including alpha divergence \cite{li-nips2016-alpha}. 

These methods also suffer from scalability issues due to the need to compute the log likelihood and its gradient over the entire dataset. Similar to the stochastic gradient version of MCMC methods introduced earlier, advances in the past decade have led to more scalable methods in this family as well, including stochastic variational inference \cite{hoffman2013stochastic}, that are able to accommodate large-scale datasets using mini-batch gradients. 

In addition to mean field VI, other more advanced approximations are possible, such as multiplicative normalizing flows (MNF) \cite{louizos2017multiplicative}, Bayesian hypernetworks \cite{Krueger2017BayesianH}, and Rank-1 factorization \cite{dusenberry2020efficient}. In multiplicative normalizing flows, we choose a simple density function such as the isotropic Gaussian distribution, and use a bijective function to transform the samples drawn from the simple density function to form more complex distributions. The Bayesian hypernetworks approach builds upon the MNF approach and uses a neural network model to transform the samples drawn from a simpler density function to model a more complex posterior distribution. Finally, the rank-1 factorization approach represents the model parameters as a product of two rank-1 factors thereby reducing the dimensionality of the base distribution of the approximate posterior. For example, if we have a parameter matrix $\pazocal{W} \in \mathbb{R}^{M \times N}$, this can be re-written as a matrix product of $W_1 \in \mathbb{R}^{M \times 1}$ and $W_2 \in \mathbb{R}^{1 \times N}$. This effectively reduces the number of parameters from $\pazocal{O}(M \cdot N)$ to $\pazocal{O}(M + N)$.

MC Dropout is a particularly interesting approach, which is equivalent to approximate variational inference under specific assumptions \cite{gal2016dropout}. Dropout itself was first introduced as a regularization technique where during every training iteration a pre-determined proportion of activations is randomly set to zero to reduce overfitting. At test time dropout is switched off and all units participate in making predictions \cite{srivastava2014dropout}. In MC Dropout, by contrast, dropout is used at prediction time. This leads to a stochastic forward pass through the point-estimated model. Multiple forward passes through the model are used and the predictive distributions are averaged. This procedure is equivalent to sampling from a specific approximate variational posterior, but has the advantage that it is very easy to implement.

The major drawback of surrogate density methods is that they introduce bias into the estimation of the posterior distribution, unless the true posterior belongs to the family of auxiliary distributions. The degree of bias will depend on the functional form of the auxiliary distribution, and the divergence measure used to estimate the parameters of the auxiliary distribution. 

In addition, under these methods, we still face hurdles in computing posterior expectations including the approximate posterior predictive distribution. In particular, even if $q(\theta|\phi)$ is a simple parametric distribution, the expectation $\mathbb{E}_{q(\theta|\phi)}[p(y|\mathbf{x}, \theta)]$ still usually cannot be computed analytically due to the non-linearity of $p(y|\mathbf{x}, \theta)$. As a result, we have to again resort to Monte Carlo approximation, and draw samples from the approximate variational posterior. Generally, these methods trade-off the potential bias in the surrogate parameter posterior for the ability to draw independent samples once the surrogate posterior parameters have been estimated.

The deployment considerations for models derived using variational inference methods are distinct from those of MCMC methods. One of the benefits of mean field variational inference over MCMC methods is that the representation of the posterior approximation only requires twice the storage of a single model. However, for this to be useful at deployment time, it must be possible to efficiently sample models from the approximate parameter posterior on the fly. The same is true of approaches like MC Dropout that require the sampling of random masks at inference time. In the case of MC Dropout, for example, current model conversion pipelines from PyTorch \cite{pytorch_neurips2019} to TensorRT \cite{tensorrt} via ONNX \cite{bai2019} strip out stochastic dropout layers. It appears that for such approaches to have practically useful storage advantages over MCMC methods on current edge hardware, additional development may be needed to enable the automated deployment of computation graphs with stochastic elements. An alternative for hardware where the computational cost of on-the-fly generation of samples from the variational posterior is prohibitive is to pre-generate and deploy a fixed variational approximate posterior ensemble. However, in this case, all the considerations that apply to the generation of MCMC posterior ensembles will also apply. 

\subsubsection{Additional Approximate Bayesian Inference Methods}

With the emergence of Generative Adversarial Networks (GANs) \cite{goodfellow2014generative}, there has been work on learning implicit generative model representations of the parameter posterior that can be used at test time to draw an arbitrary but finite number of samples from this posterior approximation. The existing work in this area involves training GANs that can approximate the posterior distribution asserted by SGLD \cite{wang2018adversarial, henning2018approximating}. To compute the approximate posterior predictive distribution, we yet again would use Monte Carlo approximation as shown in the previous subsections. An advantage of these methods is that in theory we do not need to store posterior samples for use during inference and we can also determine the number of posterior samples to use on the fly. However, the same deployment issues noted for variational inference apply here as well in terms of the feasibility of on-the-fly sampling from an auxiliary model. In addition, this approach requires allocating additional memory for the auxiliary GAN model. 

Deep Ensembles \cite{lakshminarayanan2017simple} have also shown strong performance in terms of representing predictive distributions, and can also be considered as an approximation to Bayesian inference \cite{DBLP:conf/nips/WilsonI20}. However, deep ensembles can be expensive during training, as we need to train each model in the ensemble from scratch. To alleviate this issue, snapshot ensembles \cite{DBLP:conf/iclr/HuangLP0HW17} uses a cyclical learning rate schedule to generate more ensembles in less training time. However, these methods learn a mixture of posterior modes, which is different from an approximation to the full posterior. Regardless, Deep Ensembles  have been shown to yield better performance relative to using small numbers of samples in a traditional MCMC approach.
Deployment considerations similar to those of MCMC ensembles also apply to deep ensembles.

\subsubsection{Discussion} 
As described in this section, different methods for approximating Bayesian inference have different strengths and weaknesses in terms of their theoretical and practical properties. The first such property is the bias-variance tradeoff in approximate posterior estimation. MCMC methods provide an unbiased estimate of the posterior in theory, but this relies on convergence of the underlying Markov chains, which can be time-consuming. Mixing of the Markov chains is another important factor determining the diversity of the approximate posterior ensemble. On the other hand, we can learn surrogate posterior distributions much more efficiently using stochastic gradient descent, but they introduce bias that depends on the functional choice of the auxiliary distribution and the divergence measure used. However, while important, these factors largely impact training time and the quality of approximations, not their subsequent edge deployability.

In terms of edge deployability, a common aspect among all the methods discussed in this section is that they require the application of Monte Carlo-based approximations at test time. When $S$ samples are used, the computational complexity of making predictions is $S$ times higher than if a single instance of the base model is used. As noted previously, the use of mixed precision arithmetic can help to accelerate these computations, but a reduction from 32 bit to 16 bit representations will typically at most half the prediction latency of deployed models. This is likely to be far from sufficient when considering closing the prediction latency gap between single optimization-based models and model ensembles. 

In addition, as noted previously, MCMC methods have $O(S \cdot K)$ storage cost  where $K$ is the number of parameters in the base model and $S$ is the number of samples. By contrast, mean-field variational inference has $O(2K)$ storage cost, but there appear to be practical challenges to actually realizing the benefits of the reduced storage cost of variational methods on current edge platforms. In the next sections, we turn to possible modeling and algorithmic approaches to improving the edge scalability of approximate Bayesian inference methods in terms of both computation time and storage costs. 

%% file: model_compression.tex
\section{Model Pruning Approaches to Improving\\the Scalability of Bayesian Deep Learning}
\label{sec:model_compression_pruning}
As described in the previous section, the storage and computational scalability properties of Bayesian inference for neural network models can be a significant barrier to edge and IoT deployment, despite their potential benefits in the context of intelligent systems. In this section, we discuss model pruning and sparsification approaches that aim to reduce the storage and computation requirement for Bayesian ensembles. These approaches can be broadly divided into unstructured and structured pruning approaches, as we describe below.

\subsection{Unstructured Pruning}
Optimization-based unstructured pruning methods aim to compress neural network models by sparsifying their weight matrices. The earliest work on unstructured neural network pruning dates back to the  Optimal Brain Damage method \cite{LeCun1989OptimalBD}. In the optimal brain damage approach, the authors presented a Taylor series approximation of the objective function, and show that under the assumption that the Hessian matrix is diagonal, the weights corresponding to the second order derivatives that go to zero can be removed with little to no loss in performance. The follow-up Optimal Brain Surgeon method \cite{Hassibi1993OptimalBS} highlights that the diagonal Hessian assumption can be limiting, leading to the removal of connections that should be retained. It uses second order derivatives to decide which connections to remove, and obtains better generalization on held-out test data compared to the Optimal Brain Damage approach. 

Another line of early work in unstructured pruning removes parameters based on their magnitudes. The earliest work in this area dates back to \cite{hertz1991introduction}. Since then, magnitude-based unstructured pruning has been revisited in more detail, and it has been found that iterative pruning with fine-tuning can help prune more parameters of a neural network while retaining better predictive performance compared to one-time post-hoc pruning \cite{Han2015LearningBW}. 

In the iterative pruning with fine-tuning approach, we define a pruning rate $p$ (the percentage of weights to prune) for each pruning cycle and remove all weights with magnitudes in the bottom $p$ percentile. We then fine tune the model by optimizing the unpruned weights to a desired level of convergence. We repeat these pruning and fine-tuning cycles until we achieve the desired overall sparsity level. 

Experimental results on iterative pruning with fine-tuning have shown that for the ImageNet data set \cite{deng2009imagenet}, the authors were able to reduce the number of active parameters in the AlexNet model \cite{krizhevsky2012imagenet} by 9$\times$, and reduce the number of active parameters in VGG16 \cite{simonyan2014very} by 16$\times$. It is important to note that regularization methods (such as the application of an $\ell_2$ or $\ell_1$ penalty) can help to ensure that the magnitude of the weights that are not contributing to predictive performance are driven to zero. Building upon this, there has been work on further compressing the model parameters using quantization and Huffman coding \cite{DBLP:journals/corr/HanMD15} after pruning. There has also been additional work on weight magnitude threshold-based pruning that allows for restoring connections \cite{Guo2016DynamicNS,Jin2016TrainingSD,Han2016DSDRD}.

The Lottery Ticket Hypothesis \cite{Frankle2018TheLT} proposed an iterative pruning method that is very similar to the method of \cite{Han2015LearningBW}, which we refer to as Iterative Pruning with Rewinding. This approach differs from basic iterative pruning in that following each pruning iteration, the weights used to initialize the next iteration of the algorithm are formed by combining the pruned weights with the original random weight vector (generated during initialization), instead of the weight vector obtained at the end of the previous pruning iteration. Effectively, the active weights are rewound to their initial values at the start of each round of iterative pruning. Through this approach, the authors found that there are sparse substructures within deep neural networks that, when initialized randomly, achieve very similar performance to the original dense networks with no pruning.

While unstructured pruning methods are well-studied in the context of optimization-based deep learning, they have not received as much attention in the Bayesian deep learning literature despite their potential to reduce the storage complexity of posterior model ensembles. Several applications of unstructured sparsity are possible and deserve future study. First, MCMC methods can be used to generate a posterior ensemble consisting of a set of models. A single round of pruning can then be applied to each of the models to remove a specified percentage of the smallest weights, resulting in an ensemble of weight-sparse models. However, optimization-based fine-tuning can not be applied in this setting without the potential for significantly altering the distribution that the ensemble represents. Importantly, there is potential for the sparsified models in such an ensemble to tolerate much higher levels of sparsity than an individual model due to the averaging that occurs over the elements of the ensemble when making predictions. 

An alternative approach is to use optimization-based iterative pruning and fine-tuning to derive a sparse network structure and a starting set of parameter values. MCMC methods can then be initialized to sample within this sparse structure, starting from the parameter values found using optimization. Optimization-based iterative pruning approaches can also potentially be combined with variational Bayesian deep learning methods. In the case of the classical Gaussian mean field approximation, both weights that are close to zero and weights that have high posterior variance could potentially be pruned from the model. Since variational inference is fundamentally an optimization-based procedure, it can also be composed with an iterative pruning and fine-tuning process to more closely mimic the process that is used in standard optimization-based deep learning, as described above. 

We note that while unstructured pruning has been proven to preserve prediction accuracy even at high levels of weight sparsity for large optimization-based models, whether and how these savings convert into practical savings in deployed systems is fairly complicated. We first consider the storage properties of weight-sparse models. In particular, the parameter matrices for a weight-sparse model must be stored in a compressed format to yield any storage benefit. 

Sparse matrices are typically stored either in \emph{coordinate list} (COO) format or \emph{compressed sparse column/row} (CSC/CSR) format. In these formats, at a high level, we store the indices and values of the non-zero elements of the matrix. For the COO format, for example, the space complexity of the resulting data structure is $\mathcal{O}(3N)$, where $N$ is the number of non-zero elements. Importantly, how we compose weight sparsity with Bayesian deep learning can have significant impact on the storage complexity of the resulting models. 

In the first approach described above, we considered independently sparsifying each element of the posterior ensemble. If we retain a total of $N$ non-zero weights for each of the $S$ models, the total storage required is $O(3NS)$. In the second approach, we considered sampling withing a fixed sparse structure. If this sparse structure has $N$ non-zero weights, the total required storage is theoretically, $O(2N + NS)$ since we only need to store the indices of the non-zero elements once. For $S \gg 2$, this reduces the required storage of the sparse ensemble by $66\%$ over standard COO format. While no current sparse matrix formats exist that support this optimization, it would be straightforward to implement from a storage perspective.     

Lastly, we note that realizing the computational savings of weight-sparse networks with GPUs and TPUs is also currently a challenge. PyTorch, for example, has sparse matrix support that is currently in beta testing. Existing libraries for edge GPU/TPU-accelerated architectures such as TensorRT currently only support highly restricted sparsity patterns. As a result, fully realizing the theoretical computational benefit of weight-sparse Bayesian deep learning approaches on edge platforms will require additional support for GPU/TPU-accelerated linear algebra operations over sparse matrices.

\subsection{Structured Pruning}
Optimization-based structured pruning methods apply similar pruning principles to those of unstructured pruning, but with the aim of pruning larger structural elements such as entire hidden unit or entire convolutional kernels. The simplest structured pruning method extend the iterative pruning approaches described above to also include removing hidden units or convolutional kernels that have no incoming connections \cite{Han2015LearningBW}. However, the induced sparsity patterns can be such that few units are actually pruned. These basic approaches can be improved by modifying the pruning criteria to prune units where the norm of their incoming weights is in the bottom $p$ percentile across all active units \cite{Li2016PruningFF}. This approach ensures that a desired number of units is pruned from the network on each round.

Another set of approaches leverage group LASSO (least absolute shrinkage and selection operator) regularization to encourage a group of incoming weights to go to zero simultaneously \cite{Zhang2018LEARNINGSS,alvarez2016learning,wen2016learning,He2017ChannelPF}. To apply this approach, we must first must partition the model parameters in the parameter vector $\phi$ into $K$ groups $\pazocal{G}_k$. The form of the regularizer is $R(\phi) = \sum_{k=1}^K  \big(\textstyle \sum_{j\in \pazocal{G}_k } \phi_j^2 \big)^{1/2}$ when using the group $\ell_1/\ell_2$ regularizer. For a feedforward model, we place all the incoming weights for each hidden unit into a group. Similarly, we collect all the incoming weights for a particular channel in a convolution layer together into a group. The regularizer will then tend to encourage all the weights in a group to go to zero or most of the weights in a group to be non-zero. This can make it much easier to identify structures for pruning using weight norm-based criteria and can require less fine-tuning as most inputs to a unit will tend to already be close to zero prior to pruning.

Structured pruning can also be composed with approximate Bayesian deep learning methods in multiple ways. As with unstructured pruning, we can also separately apply structured pruning to each element of a posterior ensemble. However, the chances that all incoming weights for a unit will be close to zero is many times smaller than the chance that a single weight will be small. In this case, the use of an explicit sparsity inducing prior, such as a spike-and-slab prior, would very likely be necessary to obtain meaningful pruning. The approach of using optimization-based structured pruning methods to select a structure and an initial set of weights to run MCMC-based methods from is also applicable and is a potentially promising approach. Again, there are close relationships between structured sparsity methods for optimization-based deep learning and variational Bayesian deep learning. There is also specific prior work on sparsity inducing priors for variational Bayesian methods. Previous work in this area includes the use of horseshoe priors \cite{carvalho2009handling} for approximate Bayesian inference \cite{ghosh2017model, Louizos2017BayesianCF}.

Finally, we note that there is a significant gap between the practical implementation of models that result from structured sparsity methods and unstructured sparsity methods. This is due to the fact that structured  sparsity  methods learn more compact dense models that do not need sparse matrix support to realize storage and computational savings. This is a significant advantage considering the current state of sparse matrix support in software frameworks for edge platforms. However, structured sparsity methods have been observed to require more non-zero weights than weight-sparse models to obtain similar levels of predictive performance \cite{blalock2020state}. On the other hand, the run time of linear algebra operations for sparse matrices on real hardware can have significantly more overhead than when operating over dense matrices. The trade-offs between the theoretical and practical properties of these approaches requires further study, and the best approach to use is likely to be highly dependent on the hardware and software support available on a particular deployment platform.

%% file: distillation.tex
\section{Model Distillation Approaches to Improving\\the Scalability of Bayesian Deep Learning}
\label{sec:model_compression_distillation}

In this section we describe posterior distillation methods, which provide an alternative to pruning methods that can also both decrease the storage cost and the computational cost of applying Bayesian deep learning methods. Examples of methods in this area include Bayesian Dark Knowledge (BDK) \cite{balan2015bayesian} and Generalized Posterior Expectation Distillation (GPED) \cite{Vadera2020GeneralizedBP}. These methods aim to  compress the computation of expectations under the model posterior into neural networks whose storage and computational complexity can be set to match deployment constraints. As a result, distillation approaches can expose a flexible trade-off between resource use and posterior approximation accuracy.

In the case of BDK, the selected posterior expectation is the  posterior predictive distribution. BDK approximates the posterior predictive distribution by learning an auxiliary neural network model to compress the Monte Carlo approximation to the posterior predictive distribution $\mathbb{E}_{p_{MC}(\theta|\mathcal{D},\lambda)}[p(y|\mathbf{x}, \theta)]$. The GPED approach extends the BDK approach to the case of distilling arbitrary posterior expectations. GPED has been used to directly approximate model and data uncertainty via posterior expectation distillation. 

The  major advantage of this family of  approaches is that they can drastically reduce the deployment time computational complexity of posterior predictive inference relative to using a Monte Carlo average computed using many samples. However, a shortcoming of this family of approaches is that they only capture the target posterior expectations. Thus, they do not have the ability to compute other statistics without being re-trained.

Ensemble distribution distillation (EnD\textsuperscript{2}) is a closely related approach that aims to distill the collective predictive distribution outputs of the models in an ensemble into a neural network that predicts the parameters of a Dirichlet distribution \cite{malinin2020ensemble}. The goal is to preserve more information about the distribution of outputs of the ensemble in such a way that multiple statistics of the ensemble's outputs can be efficiently approximated. We note that the EnD\textsuperscript{2} approach can be applied to any ensemble of models producing categorical output distributions and can thus be applied to distill the predictive distributions of the elements of a posterior ensemble obtained using MCMC methods as well as those obtained from a variational approximation. We also note that this approach can be extended to approximate the distribution of other posterior quantities by distilling in to approximating models that output other types of distributions. 

There is an interesting trade-off between distilling the full parameter posterior distribution into models that predict specific posterior expectations, such as BDK and GPED, and approaches that distill aspects of the posterior into distributions. As noted above, expectation distillation is a less general approach and models must be re-trained to extend coverage to additional expectations. On the other hand, EnD\textsuperscript{2} gains generality by predicting parametric distributions, from which a wider range of posterior properties can be computed. The drawback of this approach is that it introduces irreducible bias via the selection of a particular approximating family of distributions in  a way that is similar to variational inference. As a result, although a wider range of posterior properties can be estimated using EnD\textsuperscript{2}, their accuracy can be varied and all can be biased if the approximating family is not a good match to the true posterior. 

In terms of deployment on edge hardware,  one of the distinct advantages of distillation-based approximations over both MCMC and variational methods is that the deployment time computational complexity of approximating posterior expectations can be completely controlled via the selection of the model architecture that the posterior is distilled into. Once this architecture is selected and the distillation process is carried out, the result is a single model (or one model per posterior expectation of interest) that can then be deployed. This can be much simpler than deploying a posterior ensemble in the case of MCMC methods. 

However, maximizing the performance of distillation-based methods for a given computational and storage budget requires performing an architecture search to determine the optimal architecture for the approximating model. Since distillation-based learning is itself an optimization problem, one method to perform this search is to start with a large model architecture and apply iterative pruning and fine-tuning as described in the previous section. The use of both weight-level and the structure-level pruning is possible, with the deployment caveats noted in the previous section. 

The GPED paper \cite{Vadera2020GeneralizedBP} uses such an approach to expose  storage-performance and computation-performance Pareto frontiers by varying the level of pruning used. That work shows that pruning-based selection of model architectures is superior to more basic approaches like searching the space of layer-width multipliers. More generally, it demonstrates the potential for significant sensitivity to the architecture of the approximating model. The question of how to most efficiently search for an optimal distillation architecture remains an open question.

%Finally, as we conclude the review of posterior compression methods, we also note that it is certainly possible to combine the different approaches for compressing the model posterior representation. For example, we can build approaches that look at distillation plus pruning. Indeed, there is some existing that looks at this \cite{Vadera2020GeneralizedBP}. GPED approach presents an approach on posterior expectation distillation, followed by structured pruning and fine-tuning. This approach allows users to further compress the posterior representation by compressing the student model, and exposes a more performance-rich pareto frontier as compared to the width-multiplier approach to tweaking the student model capacity. This approach can also be extended to Stochastic Weight Averaging (SWA) \cite{DBLP:conf/uai/IzmailovPGVW18} by pruning and fine-tuning the final average of parameters. 

%% file: conclusions.tex
\section{Conclusions and Discussion}

In this paper we have provided a comprehensive overview of approaches to approximate Bayesian inference from the specific perspective of the challenges that emerge when considering the deployment of Bayesian deep learning models on edge hardware. We have presented a number of potential research directions aimed at improving the deployment scalability of Bayesian deep learning methods, with a focus on pruning and distillation-based model compression methods. Realizing such improvements will be crucial to enabling the practical use of Bayesian deep learning methods on edge hardware and to the development of intelligent IoT systems.

Lastly, we note that the nature of current edge hardware is likely to force trade-offs between multiple facets of the performance of Bayesian deep learning models deployed at the edge, including predictive performance, uncertainty quantification ability, robustness, and prediction latency. Methods like distillation and pruning can provide flexible trade-offs between these facets of performance, and it is imperative that candidate models are comprehensively and simultaneously evaluated with respect to all facets of performance to ensure that gains with respect to one subset of facets does not come at an unacceptable cost with respect to other subsets \cite{Vadera2020URSABenchCB}. 

It is also worth emphasizing that the approximation approaches that yield optimal Bayesian deep learning deployments in respect to one edge platform could be far from optimal for another platform (or could even be infeasible) depending on the properties of the platform, including available storage and the speed and level of parallelism of computation. The problem of learning to predict optimal deployment configurations across platforms and tasks is itself a very interesting and important research challenge in this space.

%% file: acknowledgement.tex
\section*{Acknowledgement}
This work was partially supported by the US Army Research Laboratory under cooperative agreement W911NF-17-2-0196. The   views and   conclusions  contained  in  this  document  are  those  of  the  authors  and  should  not  be interpreted as representing the official policies,  either  expressed  or  implied,  of  the  Army  Research  Laboratory  or  the  US  government. 